%% file: main.tex
\documentclass[10pt,twocolumn,letterpaper]{article}

\usepackage{cvpr}              %

\input{preamble}

\definecolor{cvprblue}{rgb}{0.21,0.49,0.74}
\usepackage[pagebackref,breaklinks,colorlinks,citecolor=cvprblue]{hyperref}

\usepackage{enumitem}
\usepackage{amsmath}
\usepackage{adjustbox}
\usepackage{boldline}

\usepackage[dvipsnames]{xcolor}
\usepackage{caption}
\usepackage{amssymb}%
\usepackage{pifont}%
\usepackage{xspace}
\providecommand{\eg}[0]{e.g.,\@\xspace}
\providecommand{\ie}[0]{i.e.,\@\xspace}
\newcommand{\cmark}{\ding{51}}%
\newcommand{\xmark}{\ding{55}}%

\newcommand{\blue}[1]{{\color{blue}{#1}}}

\def\paperID{*****} %
\def\confName{CVPR}
\def\confYear{2024}

\title{Look, Learn and Leverage (L$^3$): Mitigating Visual-Domain Shift \\ and Discovering Intrinsic Relations via Symbolic Alignment}

\author{Hanchen Xie*$^1$
\qquad Jiageng Zhu*$^1$ \qquad  Mahyar Khayatkhoei$^1$ \qquad Jiazhi Li$^1$ \qquad Wael AbdAlmageed$^2$
\\
$^1$ University of Southern California \qquad $^2$ Clemson University
}

\begin{document}
\maketitle
\let\thefootnote\relax\footnotetext{*: Equal contribution.}
\input{sections/00_abstract}
\input{sections/01_introduction}
\input{sections/02_related_works}
\input{sections/03_method}

\input{sections/04_experiments}
\input{sections/05_conclusion}
{
    \small
    \bibliographystyle{ieeenat_fullname}
    \bibliography{main}
}
\clearpage
\input{sections/xx_appendix}

\end{document}

%% file: preamble.tex
\usepackage[dvipsnames]{xcolor}
\newcommand{\red}[1]{{\color{red}#1}}

%% file: sections/00_abstract.tex
\begin{abstract}
Modern deep learning models have demonstrated outstanding performance on discovering the underlying mechanisms when both \emph{visual appearance} and \emph{intrinsic relations} (e.g., causal structure) data are sufficient, such as Disentangled Representation Learning (DRL), Causal Representation Learning (CRL) and Visual Question Answering (VQA) methods. However,  generalization ability of these models is challenged when the visual domain shifts and the relations data is absent during finetuning. To address this challenge, we propose a novel learning framework, \emph{Look}, \emph{Learn} and \emph{Leverage} (\emph{L$^3$}), which decomposes the learning process into three distinct phases and systematically utilize the class-agnostic segmentation masks as the common symbolic space to align visual domains. Thus, a relations discovery model can be trained on the source domain, and when the visual domain shifts and the \emph{intrinsic relations} are absent, the pretrained relations discovery model can be directly reused and maintain a satisfactory performance.  Extensive performance evaluations are conducted on three different tasks: DRL, CRL and VQA, and show outstanding results on all three tasks, which reveals the advantages of \emph{L$^3$}.
\end{abstract}

%% file: sections/01_introduction.tex
\section{Introduction}
\label{sec:introduction}
Machine learning (ML) methods have demonstrated great performance on recognition tasks, such as object detection \cite{yolo,faster-rcnn,vitdet} and semantic segmentation \cite{FCN,segany,segvit}, using only \emph{visual appearance} from raw visual information (\ie image and video) that diversely describes the appearance of objects. Further, when the training data is relatively comprehensive (\ie presenting sufficient \emph{intrinsic relations}, such as causal structure \cite{causalvae}), ML methods demonstrate remarkable performance on, learning not only the \emph{visual appearance}, but more importantly discovering intrinsic relations, such as object semantics \cite{vqa_v2}, object-object interactions \cite{compositional} or causal relationships \cite{causalvae}. For example, the tasks of dynamics prediction \cite{rpin}, disentangled representation learning (DRL) \cite{weakly-disentanglement}, causal representation learning (CRL) \cite{causalvae} and visual question answering (VQA) \cite{vqa_v2} entail learning the underlying physics, disentangled representations, causal structures or multi-modal reasoning mechanisms, respectively, beyond merely recognizing the objects. However, collecting data with sufficient \emph{intrinsic relations} can be expensive and even infeasible in many domains, such as natural disaster scenes and rare disease cases in the real world, which substantially limits the models' practicality and applicability. 

\input{figures-latex/concept}

Ideally, a model with good generalization capabilities applied in a new visual domain (different than that from which data was used to train the model) should perform well if the underlying relations and mechanisms stay the same. However, it is well known that existing deep learning models do not generalize well, and the model can learn to take shortcuts or encode spuriously correlated features to minimize the empirical loss \cite{mpi3d,medical_cross_domain,vdp-emc,pos_embed} so that the performance of the pretrained models rapidly decrease upon visual domain shifts. For example, the performance of disentangled representation learning methods rapidly drops when the visual domain shifts from \emph{real} to \emph{toy}, even though underlying relations and mechanism are the same \cite{mpi3d}. Further, finetuning is not always feasible due to lack of knowledge and/or data about \emph{intrinsic relations}, \eg model can not learn physics mechanism from a single image alone because a temporal sequence is usually necessary for inferring objects dynamics \cite{vdp-emc}.

To address these generalization challenges, we propose a novel learning framework, \emph{Look}, \emph{Learn} and \emph{Leverage} (\emph{L$^3$}), which adopts a \textit{two-stage} strategy (\emph{source-train} and \emph{target-adapt}) and decomposes the entire process to three phases with distinct emphasis (\emph{Look}, \emph{Learn} and \emph{Leverage}), as illustrated in \Cref{fig:concept}. \emph{Look} extracts information from (training) domain-specific \emph{visual appearance} and maps this information to a common symbolic space. \emph{Learn} discovers the \emph{intrinsic relations} and mechanisms with respect to various tasks, \eg causal mechanism for causal representation learning. Finally, \emph{Leverage} uses the pretrained relations discovery module with the corresponding \emph{Look} module trained/adjusted with only \emph{visual appearance} data in the new domain.

Inspired by the literature from dynamics prediction \cite{vdp-emc}, where the \emph{visual appearance} and the physics mechanism are learned separately, we argue that there exists a common symbolic space that is independent of the differences of the \emph{visual appearance} of various domains. This symbolic space contains information about \emph{intrinsic relations} that pertain to specific learning task. Thus, two distinct modules can be employed, which separately focus on mapping raw inputs to the common symbolic space (\emph{Look}) and learn to discover \emph{intrinsic relations} from the common symbolic space (\emph{Learn}), respectively. To this end, various \emph{visual appearance} modules can be obtained in various domains by utilizing domain-specific \emph{visual appearance} data, and the pretrained relations discovery module can be simply employed without the necessity of any infeasible adaptation that requires data from which the model can learn \emph{intrinsic relations} (e.g., temporal sequence for discovering dynamics relations) but are absent in the target domain (\emph{hence, Leverage}).

To realize the proposed \emph{L$^3$}, without the loss of generality, we use segmentation masks (SegMasks) as an example of the common symbolic space because SegMasks can retain the original spatial information while removing domain-specific visual details, which properly fits the relations discovery tasks adopted in this work (DRL, CRL and VQA). Image segmentation has lately been an active research area with varying levels of supervision \cite{iic,semi_seg,segany} in various visual domains \cite{cocodataset,gta_dataset,medi_seg}. Specifically, in this work, we use class-agnostic SegMasks extracted by a foundation model (\eg segment anything \cite{segany}) 
 to validate the proposed framework and further relax data assumptions. In contrast to conventional semantic segmentation where each mask is assigned a semantic label, class-agnostic SegMasks do not have specific labels, as illustrated in \Cref{fig:concept}. Further, an entire object can be represented by the combination of multiple segments/masks. Although class-agnostic SegMasks are relatively more general and easy to obtain, the fragmentation can increase the difficulty of discovering \emph{intrinsic relations}. We intentionally expose \emph{L$^3$} to this challenging data assumption to demonstrate the feasibility, generality, and robustness of the proposed framework even under a challenging scenario.

 Under this challenging data assumption scenario, to utilize the unstructured SegMasks systematically, we propose \emph{Mask Self-Attention Fusion (MSF)} module to self-organize all the fragmented SegMasks. Specifically, we adopt the self-attention mechanism \cite{attension} to analyze the relationships between and fuse the information of different masks. Further, depending on the task and scenario, there can be extra available data (e.g., multiple symbolic spaces and/or visual data) that can serve as auxiliary data to enhance or expand the spectrum of \emph{intrinsic relations} discovery. When auxiliary data is domain invariant, conventional feature fusion methods may be applied without the domain shift concerns. However, such an ideal case is not always feasible where the auxiliary data may be domain-specific. Thus, we design \emph{L$^3$} framework to be able to systematically leverage features from primary symbolic space and any possible auxiliary data that can even be domain-specific. To intentionally keep a relaxed data assumption, we use raw visual input (RawVis) as the example auxiliary data. Specifically, we propose \emph{Multi-Modal Cross-Attention Fusion (MMCF)} module, which utilizes the cross-attention mechanism \cite{attension} to systematically identify and integrate any extra information from auxiliary data into SegMasks features. Further, to handle the challenge that \emph{MMCF} may suffer from domain shift of the auxiliary data (\eg RawVis), we propose a novel auxiliary feature alignment strategy in the \emph{Leverage} phase, which calibrates the auxiliary feature distribution by using the primary symbolic space as an anchor. Thus, the \emph{MMCF} can produce the same feature to directly exploit the pretrained relations discovery module without extra adaptation on the absent \emph{intrinsic relations}.

 To reveal the advantages of \emph{L$^3$}, we conduct extensive studies on three tasks: Disentangled Representation Learning, Causal Representation Learning and Visual Question Answering. Our \emph{Look}, \emph{Learn} and \emph{Leverage} framework shows outstanding performance on all tasks. The contributions of this paper are:
\begin{itemize}[noitemsep,topsep=0pt,parsep=0pt,partopsep=0pt, leftmargin=*]
    \item A novel \textit{two-stage} framework, \emph{L$^3$}, that addresses the visual domain shift challenge when directly employing a pretrained \emph{intrinsic relations} discovery model on novel domain via symbolic alignment. 
    \item \emph{Mask Self-Attention Fusion} module to systematically organize the class-agnostic segmentation masks as the common symbolic space, and \emph{Multi-Modal Cross-Attention Fusion} module to further assemble the final feature with raw vision input as auxiliary.
    \item A novel alignment strategy that adapt \emph{Look} module on target domain without \emph{intrinsic relations}.
    \item Comprehensive evaluations on three different tasks (DRL, CRL and VQA) demonstrating the advantages of \emph{L$^3$}, including detailed discussions regarding the design choices. 
\end{itemize}

%% file: figures-latex/concept.tex
\begin{figure*}
    \centering
     \includegraphics[width=0.98\linewidth]{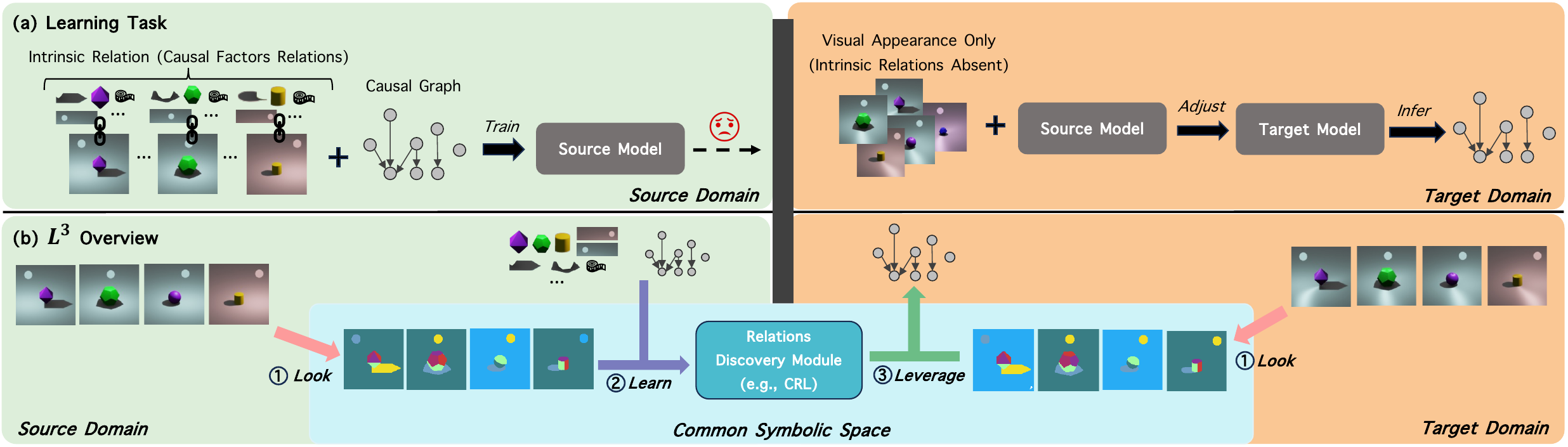}
    \caption{Conventional \emph{intrinsic relations} discovery models suffer from visual-domain shift challenge, and they can not train without \emph{intrinsic relations}. The proposed framework, \emph{Look}, \emph{Learn} and \emph{Leverage} (\emph{L$^3$}), seeks to address the challenge via symbolic alignment. \emph{Look} phase maps raw visual input from source domain to a common symbolic space where the relations discovery module is \emph{Learned}. Then, the pretrained module can be \emph{Leveraged} on the target domain with the respective \emph{Look} phase.}
    \label{fig:concept}
\end{figure*}

%% file: sections/02_related_works.tex
\section{Related Work}
\label{sec:related_works}
\textbf{Intrinsic Relations Discovery:} \emph{Intrinsic relations} discovery aims to induce correct prediction according to specific downstream tasks. Here, we summarize three different \emph{intrinsic relations} discovery tasks: disentangled representation learning (DRL), causal representation learning (CRL) and visual question answering (VQA). DRL focuses on encoding high-dimensional inputs into mutually independent latent factors, where solely utilizing \emph{visual appearance} information, i.e., unsupervised learning, was proved to be unachievable \cite{locatello2019challenging}. Thus, AdaVAE \cite{gta_dataset} was proposed for DRL by incorporating \emph{intrinsic relations} information inherent in a pair of inputs, where one sample is strategically generated by altering a certain number of generative factors according to another sample in the pair. CRL aims to simultaneously encode high-dimensional input into semantic meaningful latent factors and discover the causal structure with respect to those factors. To achieve this goal, CausalVAE \cite{causalvae} was proposed, which utilizes ground-truth generative factors as \emph{intrinsic relations} for CRL. VQA \cite{vqa_v2,mcan,simvqa} aims to answer questions regarding a visual input, which requires models to analyze \emph{intrinsic relations} that connect the visual and text modalities according to the given visual, question and answer set. 

\textbf{Domain Adaptation:} Domain adaptation in ML focuses on transferring a model trained on one visual domain to another with limited labeled data on the target domain \cite{domain-adaptation-introduce}. To achieve this, many domain adaptation methods assume the source data or at least \emph{intrinsic relations} data is available for the whole training process. Specifically, DANN \cite{DANN} was proposed for source available domain adaptation by incorporating a gradient reversal layer to enforce the feature extractor to learn a domain invariant feature. In contrast, ADDA \cite{ADDA} first trains the entire model in the source domain and adapts the feature extractor in the second stage through a domain discriminator, where both source and target data are utilized during adaptation.

%% file: sections/03_method.tex
\section{Look, Learn and Leverage (L$^3$)}
\label{sec:method}
In this section, we introduce \emph{L$^3$}, which utilizes class-agnostic segmentation masks (SegMasks) as the common symbolic space to mitigate visual-domain shift and discover \emph{intrinsic relations}. \emph{L$^3$} can be adapted and generalized to new visual domains where both the source data and \emph{intrinsic relations} on the target domain are absent. We present the problem formulation in \cref{sec:task-formulation}, the \emph{L$^3$} framework in \cref{sec:framework}, feature fusion details in \cref{sec:information-fusion} and usages of \emph{L$^3$} in example downstream tasks in \cref{sec:task-application}.

\input{figures-latex/architecture}

\subsection{Problem Formulation}
\label{sec:task-formulation}
As discussed in \cref{sec:introduction}, the learning problem in this work consists of two stages:  \emph{source-train} and \emph{target-adapt}. During the \emph{source-train} stage, the source domain dataset $\mathcal{D}_{s}$ contains both \emph{visual appearance} inputs $X_{s}$ and 
\emph{intrinsic relations}~$Y_s$, \ie $\mathcal{D}_{s} = \{X_{s}, Y_s\}$. For different tasks, \emph{intrinsic relations} $Y$ can be different, \eg generative factors in CRL or question-answer pairs in VQA. In the \emph{target-adapt} stage, the target dataset $\mathcal{D}_{t}$ only contains \emph{visual appearance} inputs $X_{t}$, \ie $\mathcal{D}_{t}= \{ X_{t}, \emptyset \}$.
Further, the source domain dataset $\mathcal{D}_{s}$ is also absent in the \emph{target-adapt} stage. 

Note that, although domain adaptation methods also aim to adapt the model to a new visual domain, these methods usually assume either source-domain data is available or both \emph{visual appearance} and \emph{intrinsic relations} data on the target domain are accessible \cite{domain-adaptation-introduce,DANN, ADDA}. Our problem formulation attempts to address a more challenging task where both source data and \emph{intrinsic relations} on the target domain are not available. Detailed discussions and studies are included in \cref{sec:discussion} and Appendix.

\subsection{L$^3$ Framework}
\label{sec:framework}
We introduce a novel learning framework, \emph{Look, Learn and Leverage} (\emph{L$^3$}) as shown in \Cref{fig:framework}, which focuses on mitigating the visual-domain shift via symbolic alignment. We adopt SegMasks as the common symbolic space because the SegMasks preserve the spatial information. To further increase the generality of \emph{L${^3}$}, we use Segment Anything (SAM)\cite{segany}, without any image-specific prompts, to generate class-agnostic SegMasks.
Further, as briefly introduced in \Cref{sec:introduction}, \emph{L${^3}$} is also capable of systematically integrating the primary symbolic space features with any auxiliary data to enhance or expand the spectrum of \emph{intrinsic relations} discovery. We use raw visual input (RawVis), e.g., images, as the example auxiliary data to show the effectiveness and generality of \emph{L{$^3$}} design. We want to highlight that, even without such auxiliary data, as will be discussed in ~\cref{sec:discussion}, \emph{L$^3$} still achieves outstanding performance on discovering \emph{intrinsic relations} that are solely presented in SegMasks. 

\subsubsection{Look: Symbolic Feature Preparation}
\label{sec:look}
We discuss \emph{L$^3$} in the order of \emph{Look, Learn and Leverage}. 
The \emph{Look} phase focuses on preparing features for the downstream task by extracting and fusing features from SegMasks and RawVis. As shown in \Cref{fig:framework}(a), given $(x_s^i, y_s^i)$ where $x_{s}^i \in X_s, y_s^i \in Y_s$, we first extract the class-agnostic SegMasks as: 
\begin{equation}
    m_s^i = F_{SAM}(x_s^i), \ \ m_s^i \in \mathbb{R}^{c \times h_m \times w_m}
\end{equation}

where we do not provide any image-specific prompts to SAM to keep a relaxed data assumption. Then, $m_s^i=\{m_{s_1}^i...m_{s_c}^i\}$ is fed to a SegMasks encoder to map each single SegMask in $m_s^i$ to the respective mask feature $u_s^i$ in a channel-wise manner. To minimize the information loss in $u_s^i$ and to facilitate downstream tasks, we reconstruct the SegMasks from $u_s^i$. This process is as follows:
\begin{equation}
\begin{gathered}
    u_s^i = F_{SegEnc}(m_s^i), \ \ u_s^i \in \mathbb{R}^{c \times h_u \times w_u}; \\
    \hat{m}_s^i = F_{SegDec}(u_s^i), \ \ \hat{m}_s^i \in \mathbb{R}^{c \times h_m \times w_m}.
\end{gathered}
\end{equation}
Because of the channel-wise processing manner, each channel in the mask feature $u_s^i$  represents the respective channel in the original SegMask $m_s^i$. As shown in \Cref{fig:mpi3d-visualization}, due to the class-agnostic nature, the combination of multiple arbitrarily ordered SegMasks may collectively represent a single semantically meaningful object. Thus, to systematically identify, connect and fuse the arbitrarily ordered mask features within $u_s^i$, we introduce the \emph{Mask Self-Attention Fusion (MSF)} module, which uses a self-attention mechanism to automatically fuse $u_s^i$ to a well-structured mask feature $u_{sf}^i$ as
\begin{equation}
    u_{sf}^i = F_{MSF}(u_s^i), \ \ u_{sf}^i \in \mathbb{R}^{c \times h_u \times w_u}.
\end{equation}
Besides SegMasks, RawVis also requires feature extraction to serve as an auxiliary. As shown in \Cref{fig:framework}(a), an RawVis $x_s^i$ is encoded to visual feature $v_s^i$ which is then used for reconstruction $\hat{x}_s^i$ by
\begin{equation}
    v_s^i = F_{VisEnc}(x_s^i); \ \ \hat{x}_s^i = F_{VisDec}(v_s^i).
\end{equation}
$F_{VisEnc}$ and $F_{VisDec}$ are autoencoders (AE) \cite{autoencoder}. Then, to assemble the information from SegMasks and RawVis according to the downstream tasks, we introduce \emph{Multi-Modal Cross-Attention Fusion (MMCF)} module, which uses cross-attention mechanism to fuse $v_s^i$ and $u_{sf}^i$ for a final feature $a_s^i$ as
\begin{equation}
    a_s^i = F_{MMCF}(v_s^i, u_{sf}^i)
\end{equation}
Details of \emph{MSF} and \emph{MMCF} are discussed in \Cref{sec:information-fusion,sec:discussion}. In addition, features from RawVis may still be vulnerable to visual-domain shift and lead to degenerate output of \emph{MMCF}. Since the RawVis only serves as auxiliary for aiding SegMasks, where SegMasks is a common symbolic space and robust to visual-domain shift, we propose to align the features between two paths so that SegMasks features can serve as an anchor for calibrating RawVis features. Since the information encoded in $u_{sf}^i$ and $v_s^i$ is different, we introduce a multilayer perceptron network, $F_{MP}$, to predict $\hat{u}_{sf}^i$ from $v_s^i$ as
\begin{equation} \label{eq:mp-train}
    \hat{u}_{sf}^i = F_{MP}(v_s^i), \ \ \hat{u}_{sf}^i \in \mathbb{R}^{c \times h_u \times w_u} 
\end{equation}
where the learned $F_{MP}$ will be reused in \emph{Leverage} phase (\Cref{sec:leverage}). In summary, \emph{Look} phase objective is as follow, where $L_{D}$ can be any distance function (e.g., mean-square error or binary cross-entropy). 
\begin{equation}
    L_{Look} = L_{D}(m_s^i, \hat{m}_s^i) + L_{D}(x_s^i, \hat{x}_s^i) + L_{D}(u_{sf}^i, \hat{u}_{sf}^i).
\end{equation}

\subsubsection{Learn: Downstream Task Learning}
\label{sec:learn}
The \emph{Learn} phase also takes place in the \emph{source-train} stage, which focuses on utilizing the final feature $a_s^i$ from \emph{Look} phase to train the downstream task module (\ie \emph{intrinsic relations} discovery). Further, an AE-based cross reconstruction regularization is also conducted for preserving information from both SegMasks and RawVis to facilitate relations discovery. Following, we use disentangled representation learning (DRL) as an example task for illustrating the complete \emph{Learn} phase. DRL details and the \emph{Learn} phase's application to other tasks will be discussed in \Cref{sec:task-application} and Appendix.

As shown in \Cref{fig:framework}(a), by taking $a_s^i$ as input, a  variational AE (VAE) \cite{vae} downscales $a_s^i$ to the disentangled feature $z_s$ and produces  $\hat{a}_s^i$. $z_s$, along with $y_s^i \in Y_s$, will be used as the input for the downstream task training (DRL in this example) to produce the respective task loss, $L_{Task}$. To better facilitate the task training by enriching the information in $a_s^i$, we apply an additional set of reconstructions by taking $\hat{a}_s^i$ as input to reconstruct the original SigMasks and RawVis as
\begin{equation}
    \tilde{x}_s^i = F_{XVisDec}(\hat{a}_{s}^i); \ \ \tilde{m}_s^i = F_{XSegDec}(\hat{a}_{s}^i)
\end{equation}
The overall training objective of $Learn$ phase and the entire \emph{source-train} stage can be summarised as
\begin{equation}
\begin{gathered}
    L_{Learn} = L_{Task} + L_D(\tilde{m}_s^i, m_s^i) + L_D(\tilde{x}_s^i, x_s^i) \\
    L_{source-train} = L_{Look} + L_{Learn}
\end{gathered}
\end{equation}

\subsubsection{Leverage: Relations Discovery in the Target Domain}
\label{sec:leverage}
The \emph{Leverage} phase takes place in \emph{target-adapt} stage, which employs the relations discovery module obtained in the \emph{Learn} phase while aligning RawVis feature on the target domain. When the visual domain shifts, recalling the problem formulation in \Cref{sec:task-formulation}, both target \emph{intrinsic relations} $Y_t$ and the entire source data $D_{s}$ are absent. Thus, conventional domain adaption methods are challenged. 
\input{figures-latex/mpi3d_results}
Contrastively, the common symbolic space in \emph{L${^3}$}, SegMasks, provides domain-invariant information to employ the pretrained task module as well as serves as the anchor to align the RawVis features. Thus, the main focus of the \emph{Leverage} phase, as shown in \Cref{fig:framework}(b), is to align the RawVis feature $v_t^i$ by using SegMasks feature $u_{tf}^i$ as an anchor, where $v_t^i$ and $u_{tf}^i$ are obtained by consuming $x_t^i \in X_t$ as (same with the \emph{Look} phase in \cref{sec:look}):
\begin{equation}
\begin{gathered}
    m_t^i = F_{SAM}(x_t^i); \ \ v_t^i = F_{VisEnc}(x_t^i); \\
    u_t^i = F_{SegEnc}(m_t^i); \ \ u_{tf}^i = F_{MSF}(u_t^i)
\end{gathered}
\end{equation}
Since the focus of \emph{Leverage} phase is to adjust the distribution of $v_t$ on the target domain, only $F_{VisEnc}$ and $F_{VisDec}$ are updated, and all other modules are fixed. Recalling $F_{MP}$ is prepared in \emph{Look} phase, we fix $F_{MP}$ and produce $\hat{u}_{tf}^i$ with $v_t^i$ according to \Cref{eq:mp-train}. The objective in the \emph{target-adapt} stage is:
\begin{equation}
    L_{Leverage} = L_D(\hat{u}_{tf}^i, u_{tf}^i) + L_D(\hat{x}_t^i, x_t^i)
\end{equation}
After \emph{target-adapt}, when testing on target domain, the \emph{L$^3$} intakes sample $x$ and produces the respective $z$, serving as the input of the downstream task module for relations discovery. 

\subsection{Feature Fusion}
\label{sec:information-fusion}
In \Cref{sec:framework}, we introduced the proposed \emph{L$^3$} framework, where SegMasks is used as the primary symbolic space for the downstream tasks and RawVis is used as the auxiliary data. Further, SegMasks are class-agnostic, where each mask is not associated with a semantically meaningful label, and the combination of several masks may collectively represent a single object, as shown in \Cref{fig:mpi3d-visualization}. Thus, in \Cref{sec:look}, we introduced the \emph{Mask Self-Attention Fusion} module and \emph{Multi-Modal Cross-Attention Fusion} module to organize and fuse features from both SegMasks and RawVis.

The \emph{\textbf{Mask Self-Attention Fusion (MSF)}} module uses a self-attention mechanism to automatically organize SegMasks features $u^i$ to produce $u_{f}^i$ as follow, where $Q$, $K$, and $V$ are the linear query, key, and value projectors in $F_{MSF}$, respectively:
\begin{equation}
    u_{f}^i = F_{MSF}(u^i) = \text{softmax}(\frac{Q(u^i) K^{T}(u^i)}{\sqrt{d_k}})V(u^i)
\end{equation}

The \emph{\textbf{Multi-Modal Cross Attention Fusion (MMCF)}} module aims to fuse features from SegMasks and RawVis. Specifically, we use a cross-attention mechanism to systematically supplement $u_{f}^i$ with $v^i$. Since under the design concept of \emph{L$^3$}, $u_{f}^i$ is the major information source for the downstream task due to visual domain invariant and $v^i$ serve as the auxiliary data as needed, we use $v^i$ as query and $u_{f}^i$ as the key and value. Empirical studies on this design choice are included in \Cref{sec:discussion}. Thus, by taking $v^i$ and $u_f^i$ as input, the \emph{MMCF} produce the final feature $a^i$ as follow, where $Q'$, $K'$, and $V'$ are the linear query, key, and value projectors in $F_{MMCF}$, respectively:
\begin{equation}
    u_f^i = F_{MMCF}(v^i, u_f^i) = \text{softmax}(\frac{Q'(v^i) K'^{T}(u^i_{f})} {\sqrt{d_k}})V'(u^i_{f})
\end{equation}

\subsection{Downstream Task Application}
\label{sec:task-application}
Here, we describe the applications of \emph{L$^3$} in DRL, CRL and VQA. Since downstream task modules are not our contribution, we summarize the principles of each task and refer the readers to the Appendix and original papers for details. \textbf{DRL} aims to learn to encode high-dimensional data into mutually independent latent factors, where each latent factor contains one generative factor information \cite{bengio_representaion}. We utilize AdaVAE \cite{weakly-disentanglement} as the DRL downstream task module. During training, a pair of inputs serves as the \emph{intrinsic relations} where one sample in the pair is generated by altering $k$ generative factors corresponding to another sample. 
\textbf{CRL} aims to simultaneously encode the high-dimensional data into semantically meaningful latent factors and correctly discover the causal relationships among those factors. To achieve this goal, we adapt CausalVAE \cite{causalvae} as the CRL module. Since CausalVAE is trained in a fully supervised setting, to train CausalVAE on the source domain, the ground-truth generative factors serve as the \emph{intrinsic relations}. 
\textbf{VQA} aims to analyze and integrate features from visual and text modalities, so that the model can predict the answers according to the visual and question input. Thus, we take the question and answer pairs as $Y$ where the model can only access the visual input while \emph{Leverage} alignment. Following TDW-VQA dataset \cite{simvqa}, which serves as the target domain, we use MCAN \cite{mcan} as the VQA module. In \emph{$L^3$}, a simple feed-forward network maps $a^i$ to $\hat{a}^i$ to avoid potential information bottleneck, and $\hat{a}^i$ is fed to MCAN as the VQA visual input. 

\input{tables-latex/causal_results}
\input{figures-latex/vqa_align_ablation}

%% file: figures-latex/architecture.tex
\begin{figure*}
    \centering
     \includegraphics[width=0.95\textwidth]{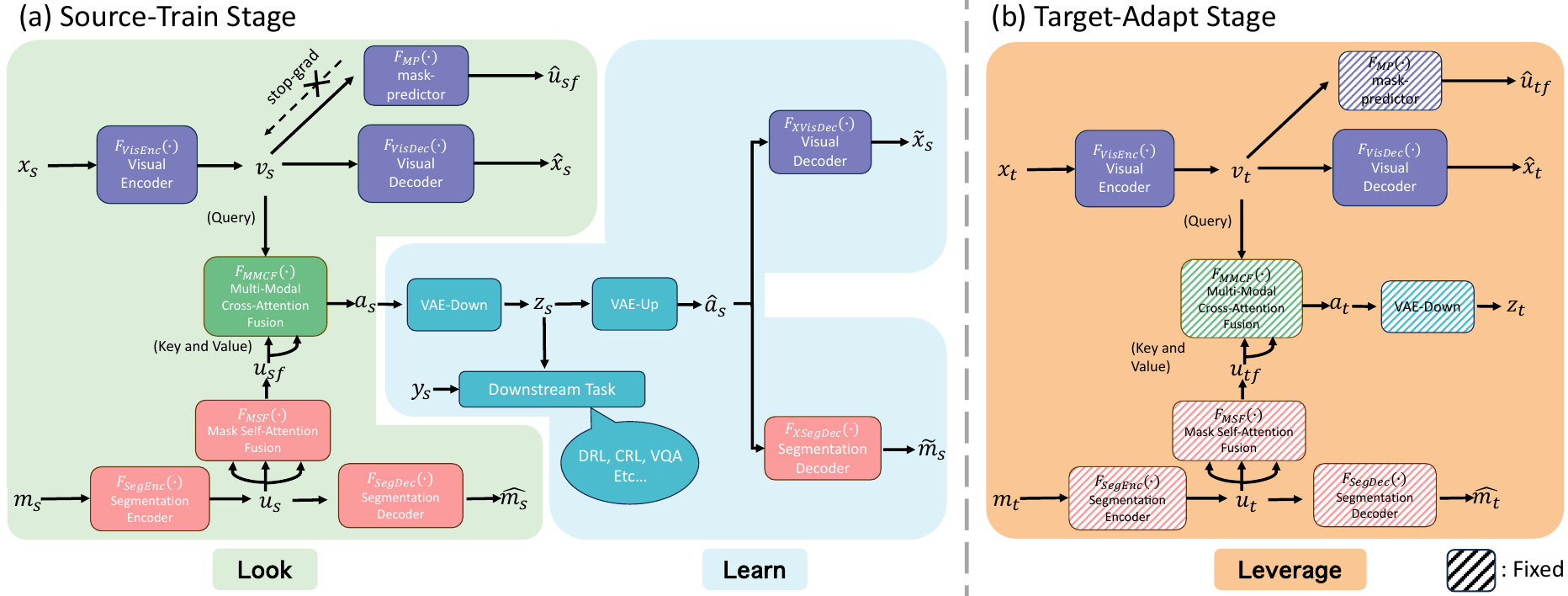}
    \caption{\emph{Look}, \emph{Learn} and \emph{Leverage} (\emph{L$^3$}) framework.} %
    \label{fig:framework}
\end{figure*}

%% file: figures-latex/mpi3d_results.tex
\begin{figure*}
    \centering
     \includegraphics[width=0.95\textwidth]{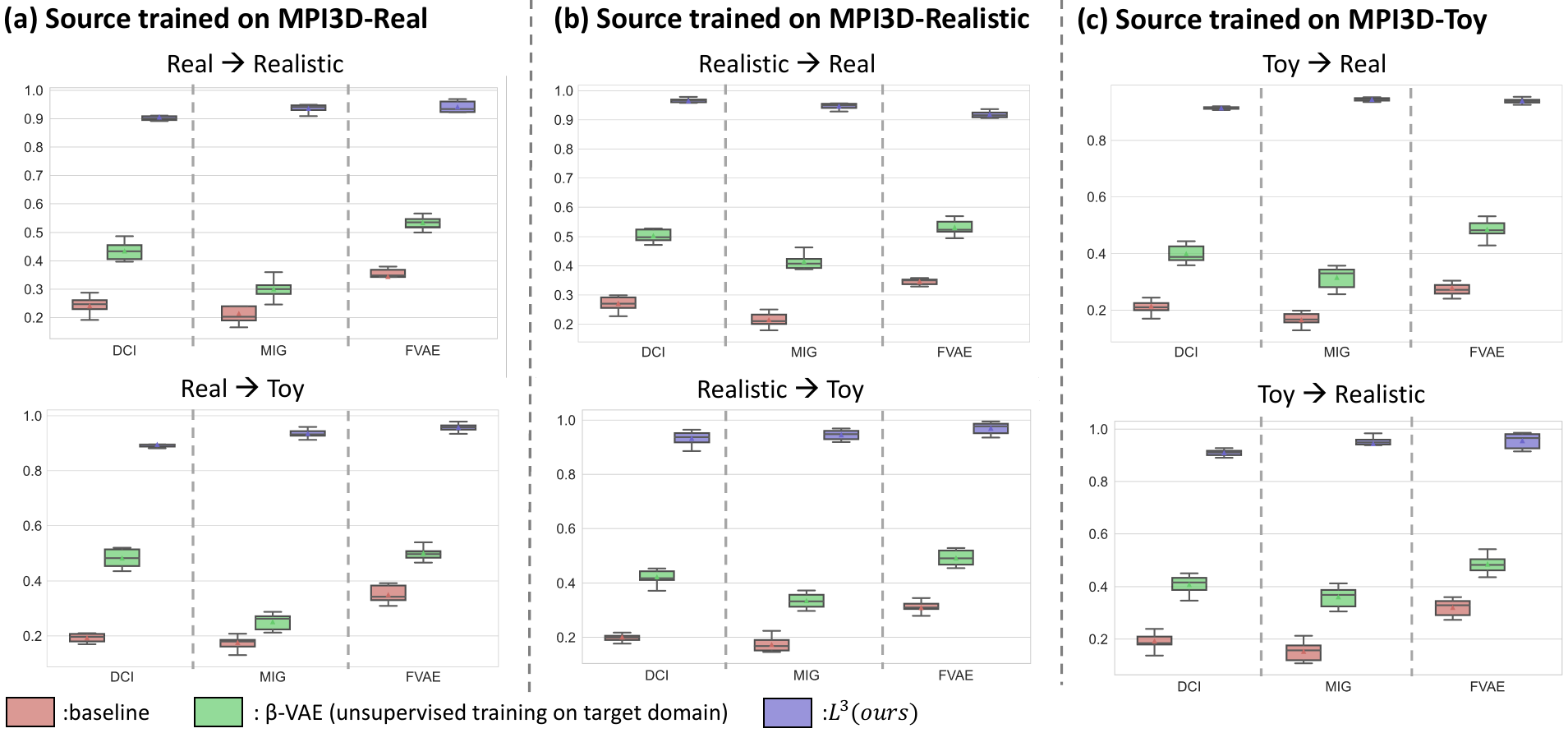}
    \caption{Performance of DRL task on MPI3D dataset. Normalized results are reported and the original results are in Appendix.} %
    \label{fig:mpi3d-total-results}
\end{figure*}

%% file: tables-latex/causal_results.tex
\begin{table*}[ht]
\centering
\caption{Performance of CRL task on the Shadow Datasets. Normalized results are reported and the original results are in Appendix.}
\label{table:causal_results}
\renewcommand{\arraystretch}{1.2}
{
\begin{adjustbox}{width=0.75\textwidth}
\setlength{\tabcolsep}{0.3em}
\begin{tabular}{ccccc|cccc}
\hlineB{3}
Dataset   & \multicolumn{4}{c|}{Shadow-Sunlight}                                      & \multicolumn{4}{c}{Shadow-Pointlight}                                    \\ \hline
Direction & \multicolumn{2}{c|}{Original $\rightarrow$ Metal}   & \multicolumn{2}{c|}{Metal $\rightarrow$ Original} & \multicolumn{2}{c|}{Original $\rightarrow$ Metal}   & \multicolumn{2}{c}{Metal $\rightarrow$ Original} \\ \hline
Models    & $F_1^{MIC}\uparrow$ & \multicolumn{1}{c|}{$F_1^{TIC}\uparrow$ } & $F_1^{MIC}\uparrow$           & $F_1^{TIC}\uparrow$          & $F_1^{MIC}\uparrow$ & \multicolumn{1}{c|}{ $F_1^{TIC}\uparrow$ } & $F_1^{MIC}\uparrow$          & $F_1^{TIC}\uparrow$          \\ \hline
Baseline  & 0.67 \tiny{$\pm$ 0.05}   & \multicolumn{1}{c|}{0.68 \tiny{$\pm$ 0.04}}   & 0.72 \tiny{$\pm$ 0.04}             & 0.74 \tiny{$\pm$ 0.03}           & 0.71 \tiny{$\pm$ 0.04}   & \multicolumn{1}{c|}{0.69 \tiny{$\pm$ 0.05}}   & 0.74 \tiny{$\pm$ 0.02}            & 0.78 \tiny{$\pm$ 0.03}           \\
UT        & 0.40 \tiny{$\pm$ 0.04}   & \multicolumn{1}{c|}{0.36 \tiny{$\pm$ 0.05}}   & 0.37 \tiny{$\pm$ 0.02}             & 0.39 \tiny{$\pm$ 0.03}            & 0.38 \tiny{$\pm$ 0.04}   & \multicolumn{1}{c|}{0.37 \tiny{$\pm$ 0.05}}   & 0.40 \tiny{$\pm$ 0.03}            & 0.42 \tiny{$\pm$ 0.04}            \\
\emph{L$^3$} (Ours)  & \textbf{0.95 \tiny{$\pm$ 0.03}}   & \multicolumn{1}{c|}{\textbf{0.97\tiny{$\pm$ 0.05}}}   & \textbf{0.94 \tiny{$\pm$ 0.02}}             & \textbf{0.92 \tiny{$\pm$ 0.03}}            & \textbf{0.95  \tiny{$\pm$ 0.02}}   & \multicolumn{1}{c|}{\textbf{0.94 \tiny{$\pm$ 0.02}}}   & \textbf{0.95 \tiny{$\pm$ 0.02}}            & \textbf{0.94 \tiny{$\pm$ 0.03}}            \\ \hlineB{3}
\end{tabular}
\end{adjustbox}
}
\end{table*}

%% file: figures-latex/vqa_align_ablation.tex
\begin{figure*}
    \centering
     \includegraphics[width=0.95\textwidth]{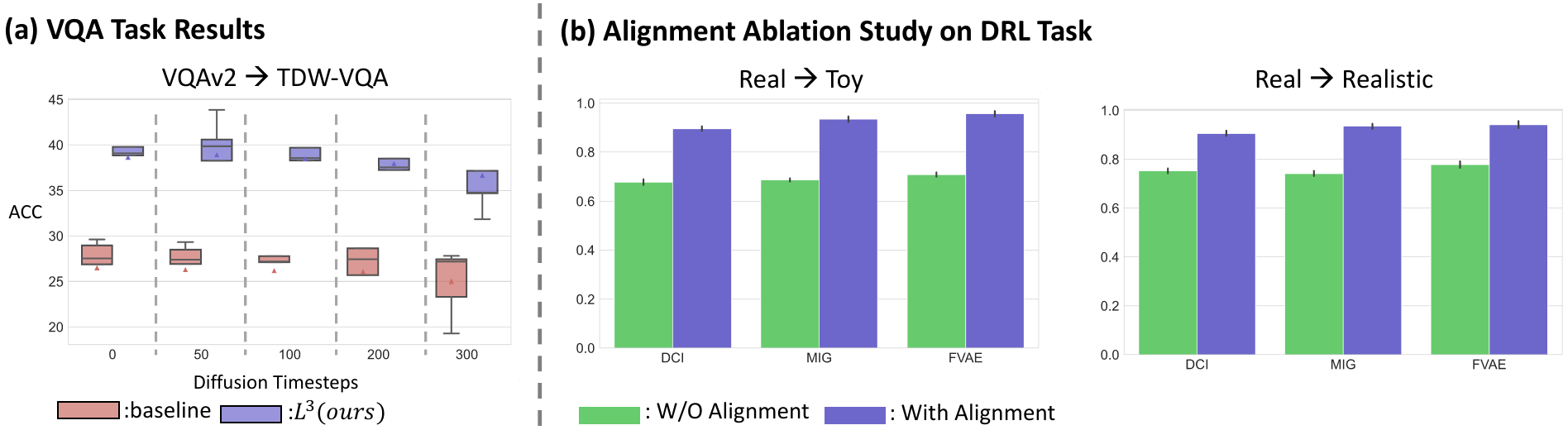}
    \caption{(a): Performance of VQA task when visual domain shifts from VQAv2 to TDW-VQA. Various diffusion timesteps are introduced to the TDW-VQA visual feature to increase the distribution shift. (b): Ablation study of feature alignment in \emph{Leverage} step; Normalized results are reported.}
    \label{fig:vqa_predictor_ablation}
\end{figure*}

%% file: sections/04_experiments.tex
\section{Experiments}
\label{sec:experiments}
\subsection{Experiments on Downstream Tasks}
\label{sec:experiment_details}
\textbf{DRL} experiments are conducted on MPI3D dataset \cite{mpi3d}, which contains three different visual domain subsets: \emph{Toy}, \emph{Realistic} and \emph{Real}. As discussed in \Cref{sec:task-application}, we select AdaVAE as the downstream task module in \emph{L$^3$} and use $\beta$-VAE \cite{betavae}, an unsupervised training (UT) method, as an additional baseline. RGB images are provided to serve as $X$ in both the source and target domain, and supervision signals are provided as \emph{intrinsic relations} ($Y_s$) only in the source domain. We utilize mutual information gap (MIG) \cite{mig}, DCI-Disentanglement (DCI) \cite{dci} and FactorVAE score (FVAE) \cite{factorvae} as evaluation metrics. The results are shown in \Cref{fig:mpi3d-total-results}.

\textbf{CRL} experiments are conducted on Shadow-Datasets \cite{shadows}, which includes two distinct datasets: \emph{Shadow-Sunlight} and \emph{Shadow-Pointlight}, where each simulates different causal mechanism. Thus, for proper evaluation, respective dataset pair in different visual domains need to be generated for both \emph{Sunlight} and \emph{Pointlight} datasets. We use the Blender file provided by \cite{shadows}, adjust the materials of objects and floor, which does not affect the causal mechanisms of interest, and create the dataset pairs: \emph{Shadow-Sunlight: Metal} and \emph{Shadow-Pointlight: Metal}. Following \cite{do-VAE}, we adopt $F_1^{MIC}$ and $F_1^{TIC}$ to evaluate the correctness and falseness of learned causal relations. We also use a UT method, $\beta$-VAE, as an additional baseline. The results are shown in \Cref{table:causal_results}. 

\input{tables-latex/segmask_only}
\input{figures-latex/mpi3d_visualization}

\textbf{VQA} experiments are conducted by pairing VQAv2 \cite{vqa_v2} and TDW-VQA \cite{simvqa} datasets, which contains real and synthetic images, respectively. We focus on counting questions to match question type. Question and answer pairs serve as \emph{intrinsic relations}. The visual features, extracted by bottom-up-attention model \cite{botton-up-att} pretrianed on visual genome \cite{vgenome}, serve as visual input which follows \cite{mcan,simvqa}. To better show the robustness of \emph{L$^3$} under potentially complicated scenarios, we follow DDIM \cite{ddim} and introduce various timesteps to visual inputs on the target domain. The results are shown in \Cref{fig:vqa_predictor_ablation}. 

More details on the datasets and downstream tasks are in Appendix.

\subsection{Comparison to Baselines}
\label{sec:compare_baseline}
We choose the original task modules, which are trained on source domain and directly applied on target domain as baselines. We normalize the DRL and CRL metrics by dividing the model score on the target domain by the respective score on the source domain. A normalized value closer to one indicates a smaller score gap between the source and target domain. This normalization emphasizes the focus on model generalization which bridges the gap for employing the pretrained relations discovery model. Also, since all model performances are comparable on the source domain, the normalized metrics can indicate both the performance gap that we aim to close and the actual performance difference between models on the target domain. All metrics values without normalization are included in the Appendix. We do not normalize the VQA results due to the complexity of the VQA task. 

As shown in \Cref{fig:mpi3d-total-results,table:causal_results,fig:vqa_predictor_ablation}, \emph{L$^3$} consistently outperforms all baselines on all three downstream tasks, which reveals its advantages. It is noticeable that on DRL and CRL, the normalized metrics of \emph{L$^3$} are all close to one, which indicates that \emph{L$^3$} can greatly mitigate the challenge and empower the relations discovery models to maintain their capacity even when the object appearance is largely altered. Further, on the complex VQA task that involves multi-modal data, \emph{L$^3$} also clearly outperforms the baseline, which reveals its great potential and robustness on sophisticated relations discovery tasks. 

\subsection{Ablation Studies and Discussion}
\label{sec:discussion}
\textbf{Feature Fusion: } \emph{MSF} and \emph{MMCF} are two critical feature fusion modules in \emph{L$^3$}. Since, as shown in \cref{fig:mpi3d-visualization}, the combination of several unstructured class-agnostics SegMasks may represent a single object, it is critical to systematically fuse SegMasks features. As shown in \Cref{table:att_ablation}, without \emph{MSF}, the arbitrarily ordered features can confuse the following processes and jeopardize the model performance. After obtaining a consistently ordered mask feature $u_{f}^i$, \emph{MMCF} is employed to fuse the $u_{f}^i$ with the auxiliary RawVis feature $v^i$ for a final feature $a^i$. A simple alternative fusion method is concatenating $u_{f}^i$ with $v^i$ and feeding it to a fully connected network. In contrast, \emph{MMCF} systematically fuses the feature where $v^i$ serves as an auxiliary data to enhance \emph{intrinsic relations} discovery. Since SegMasks is domain-invariant, it can reliably serve as the major information source, whereas the auxiliary data may be domain-specific. Thus, we use $u_f^i$ as the key and value, and $v^i$ as the query. As shown in \cref{table:att_ablation}, both simple fusion and using $u_f^i$ as query can jeopardize the \emph{leverage} performance.

\input{tables-latex/attention_domain_ablation}
\textbf{\emph{L$^3$} Without Auxiliary RawVis:} As discussed in \cref{sec:introduction}, we intentionally assume a challenging scenario to realize \emph{L$^3$} by selecting class-agnostic SegMasks as the primary symbolic space and RawVis as auxiliary data. We want to emphasize that RawVis only serves as auxiliary data to enhance or expand the \emph{intrinsic relations} spectrum that \emph{L$^3$} can discover. Thus, without RawVis, \emph{L$^3$} can still discover the relations that are solely presented by SegMasks. To validate this argument, we provide both quantitative and qualitative studies in \cref{tab:ablation_segmask_only,fig:mpi3d-visualization}, respectively. For the quantitative study, we separate the disentangled factors in DRL tasks on MPI3D dataset into two groups: SegMask-Presented group (object shape, object size, camera height, horizontal axis, and vertical axis) and Auxiliary-Expanded group (object color and background color). We separately evaluate the model performance with respect to each group. On the target domain and without RawVis, $L^3$ (w/o RawVis) still achieves outstanding DRL performance on the SegMask-Presented group and outperforms baseline models by a large margin. Those results further reveal the effectiveness and robustness of $L^3$. Further, as illustrated in \cref{fig:mpi3d-visualization}, for $L^3$ (w/o RawVis), all the SegMask-Presented factors (e.g., shape, size, etc.) are successfully reconstructed, whereas the baseline fails to correctly reconstruct factors.

\textbf{\emph{Leveraging} Alignment: } In the \emph{Leverage} task, since the distribution of RawVis feature may shift in the new visual domain, we introduce an alignment strategy by using the feature from symbolic space as an anchor. As shown in \cref{fig:vqa_predictor_ablation}, removing the alignment can lead to sub-optimal results because the unaligned $v_t^i$ can confuse \emph{MMCF} so that it can not produce a satisfactory $a_t^i$ for relations discovery. Noticeably, even without the alignment, \emph{L$^3$} can still outperform baselines in \cref{fig:mpi3d-total-results}, which further demonstrates the robustness of \emph{L$^3$} and the auxiliary role of  $v^i$. 

\textbf{Comparison to Domain Adaptation: } As mentioned in \cref{sec:task-formulation}, our problem formulation is different from conventional domain adaptation because both source data and \emph{intrinsic relations} data on the target domain are absent. To complete the discussion and to show the advantage of \emph{L$^3$}, despite being an unfair comparison for \emph{L$^3$}, we compare the DRL performance with two general source-available domain adaptation methods: DANN \cite{DANN} and ADDA \cite{ADDA}, where the results are shown in \cref{table:domain_adapt}. During the whole training and/or adapting process, even though the entire source data is available, the \emph{intrinsic relations} in target domain are absent. Without strategically leveraging the \emph{intrinsic relations} in source domain, those source-available domain adaptation methods fail to achieve satisfactory performance in the target domain compared to their performance in the source domain. In contrast, even though the source data is absent in \emph{Leverage} phase, by aligning the target visual feature $v_t^i$ with feature $u_{tf}^i$ from symbolic space through fixed $F_{MP}$, the $F_{SegEnc}$, \emph{MSF} and \emph{MMCF} serve as inductive bias \cite{locatello2019challenging}, which enforces the \emph{L$^3$} maintaining its good performance on the target domain. 

%% file: tables-latex/segmask_only.tex
\begin{table*}[ht]
\centering
\caption{DRL applied to MPI-3D demonstrating the performance of $L^3$ on the target domain without raw visual input on different disentangled factor groups. Normalized metrics are reported. }
\label{tab:ablation_segmask_only}
\renewcommand{\arraystretch}{1.2}
{
\begin{adjustbox}{width=0.98\linewidth}
\setlength{\tabcolsep}{0.15em}
\begin{tabular}{ccccccccccccc}
\hlineB{3}
                                  & \multicolumn{6}{c|}{Real $\rightarrow$ Realistic}                                                                                                       & \multicolumn{6}{c}{Real $\rightarrow$ Toy}                                                                                                    \\ \hline
                                  & \multicolumn{3}{c|}{SegMask-Presented}                                   & \multicolumn{3}{c|}{Auxiliary-Expanded}                                  & \multicolumn{3}{c|}{SegMask-Presented}                                    & \multicolumn{3}{c}{Auxiliary-Expanded}                        \\ \hline
                                  & DCI $\uparrow$             & MIG $\uparrow$              & \multicolumn{1}{c|}{F-VAE $\uparrow$}           & DCI $\uparrow$             & MIG $\uparrow$              & \multicolumn{1}{c|}{F-VAE $\uparrow$}           & DCI $\uparrow$             & MIG $\uparrow$              & \multicolumn{1}{c|}{F-VAE $\uparrow$}            & DCI $\uparrow$              & MIG $\uparrow$              & \multicolumn{1}{c}{F-VAE $\uparrow$} \\ \hline
Baseline                          & 0.23 \tiny{$\pm$ 0.02}& 0.25 \tiny{$\pm$ 0.02}& \multicolumn{1}{c|}{0.33 \tiny{$\pm$ 0.02}} & 0.18 \tiny{$\pm$ 0.01}  & 0.19 \tiny{$\pm$ 0.01}& \multicolumn{1}{c|}{0.27 \tiny{$\pm$ 0.02}}& 0.16 \tiny{$\pm$ 0.03} & 0.29 \tiny{$\pm$ 0.03}  & \multicolumn{1}{c|}{0.25 \tiny{$\pm$ 0.02}}  & 0.24 \tiny{$\pm$ 0.02}& 0.25 \tiny{$\pm$ 0.02}  & 0.24 \tiny{$\pm$ 0.01}            \\
UT                                & 0.40 \tiny{$\pm$ 0.02} & 0.26 \tiny{$\pm$ 0.02}  & \multicolumn{1}{c|}{0.35 \tiny{$\pm$ 0.02}} & 0.34 \tiny{$\pm$ 0.03}  & 0.30 \tiny{$\pm$ 0.03}  & \multicolumn{1}{c|}{0.32 \tiny{$\pm$ 0.02}} & 0.34 \tiny{$\pm$ 0.01} & 0.18 \tiny{$\pm$ 0.03}  & \multicolumn{1}{c|}{0.26 \tiny{$\pm$ 0.02}}  & 0.32 \tiny{$\pm$ 0.01}  & 0.31 \tiny{$\pm$ 0.02}  & 0.34 \tiny{$\pm$ 0.04}            \\
\emph{L$^3$}              & 0.95 \tiny{$\pm$ 0.01} & 0.94 \tiny{$\pm$ 0.002} & \multicolumn{1}{c|}{0.95 \tiny{$\pm$ 0.01}} & 0.90 \tiny{$\pm$ 0.01}  & 0.89 \tiny{$\pm$ 0.004} & \multicolumn{1}{c|}{0.93 \tiny{$\pm$ 0.01}} & 0.94 \tiny{$\pm$ 0.01} & 0.93 \tiny{$\pm$ 0.004} & \multicolumn{1}{c|}{0.91 \tiny{$\pm$ 0.003}} & 0.96 \tiny{$\pm$ 0.005} & 0.92 \tiny{$\pm$ 0.004} & 0.92 \tiny{$\pm$ 0.004}           \\
\emph{L$^3$} (w/o RawVis) & 0.93 \tiny{$\pm$ 0.01} & 0.92 \tiny{$\pm$ 0.005} & \multicolumn{1}{c|}{0.93 \tiny{$\pm$ 0.01}} & 0.24 \tiny{$\pm$ 0.003} & 0.31 \tiny{$\pm$ 0.004} & \multicolumn{1}{c|}{0.36 \tiny{$\pm$ 0.01}} & 0.92 \tiny{$\pm$ 0.01} & 0.89 \tiny{$\pm$ 0.003} & \multicolumn{1}{c|}{0.91 \tiny{$\pm$ 0.001}} & 0.24 \tiny{$\pm$ 0.02}  & 0.32 \tiny{$\pm$ 0.02}  & 0.37 \tiny{$\pm$ 0.02}            \\ \hlineB{3}
\end{tabular}
\end{adjustbox}
}
\end{table*}

%% file: figures-latex/mpi3d_visualization.tex
\begin{figure*}
    \centering
     \includegraphics[width=0.95\textwidth]{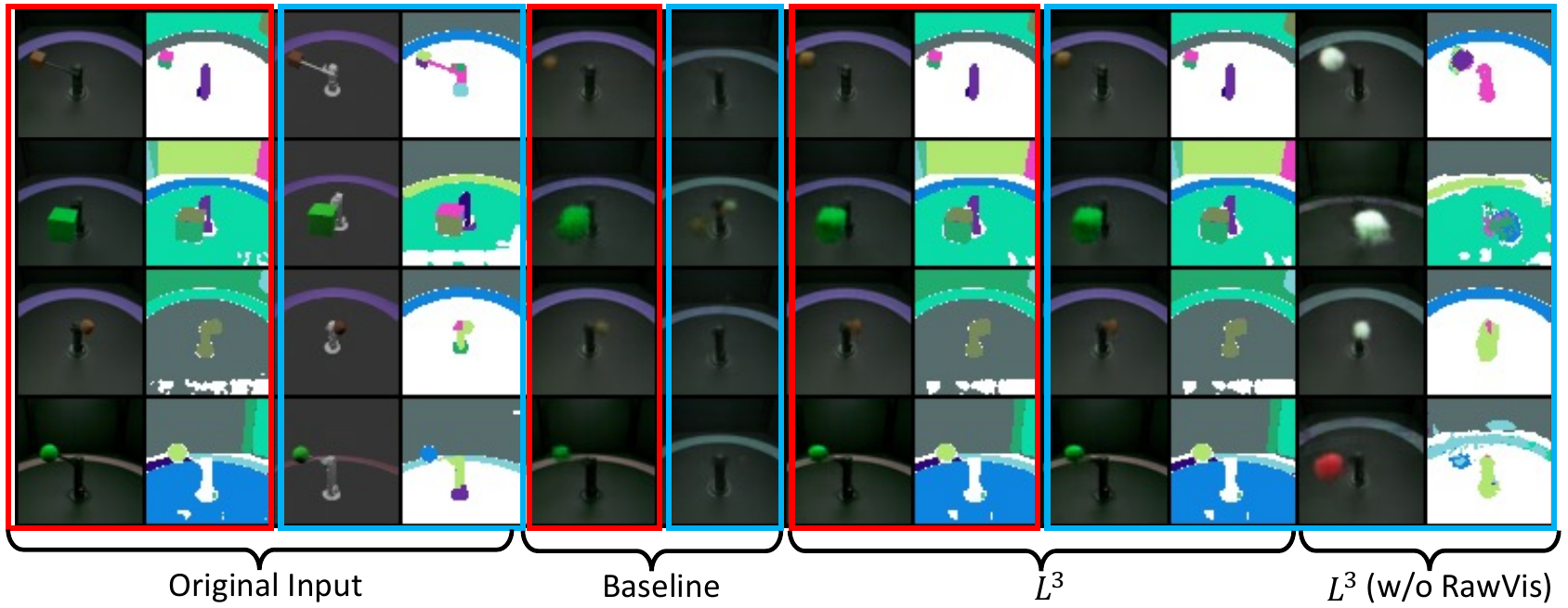}
    \caption{Visualization of $m^i$ and $x^i$ reconstruction on MPI3D dataset. Source domain (\red{red box}) is \emph{Real} and target domain (\blue{blue box}) is \emph{Toy}. Baseline fail to make any meaningful output on the target domain due to visual-domain shift, whereas \emph{L$^3$} has meaningful output on both source and target domain. \emph{L$^3$}'s outputs follow the source domain which reveals the advantage of \emph{Leverage} alignment.} %
    \label{fig:mpi3d-visualization}
\end{figure*}

%% file: tables-latex/attention_domain_ablation.tex
\begin{table*}[ht]
    \caption{Ablation study of feature fusion (left) and comparison with domain adaptation (right) on DRL task. \emph{Rev MMCF} means using $u_f^i$ as query. Normalized results are reported. }
    \label{tab:ablation_fusion_adapt}
\begin{minipage}[c]{0.49\textwidth}
\centering
\label{table:att_ablation}
\raggedright
\renewcommand{\arraystretch}{1.21}
{
\begin{adjustbox}{width=0.97\textwidth}
\setlength{\tabcolsep}{0.3em}
\begin{tabular}{cccc|ccc}
\hlineB{3}
Direction                    & \multicolumn{3}{c|}{Real $\rightarrow$ Toy}                                                                        & \multicolumn{3}{c}{Real $\rightarrow$ Realistic}                                                                  \\ \hline
Models                       & DCI $\uparrow$                              & MIG $\uparrow$                              & FVAE $\uparrow$                              & DCI $\uparrow$                              & MIG $\uparrow$                              & FVAE $\uparrow$                             \\ \hline
w/o MSF                      & 0.45 \tiny{$\pm$ 0.02}                  & 0.48 \tiny{$\pm$ 0.03}                     & 0.59 \tiny{$\pm$ 0.03}                      & 0.44 \tiny{$\pm$ 0.02}                     & 0.47 \tiny{$\pm$ 0.03}                     & 0.58 \tiny{$\pm$ 0.04}                     \\
w/o MMCF                     & 0.42 \tiny{$\pm$ 0.04}                     & 0.43 \tiny{$\pm$ 0.02}                     & 0.67 \tiny{$\pm$ 0.04}                      & 0.40 \tiny{$\pm$ 0.04}                     & 0.48 \tiny{$\pm$ 0.03}                     & 0.67 \tiny{$\pm$ 0.04}                     \\
\multicolumn{1}{l}{Rev MMCF} & \multicolumn{1}{l}{0.46 \tiny{$\pm$ 0.03}} & \multicolumn{1}{l}{0.48 \tiny{$\pm$ 0.03}} & \multicolumn{1}{l|}{0.66 \tiny{$\pm$ 0.03}} & \multicolumn{1}{l}{0.49 \tiny{$\pm$ 0.06}} & \multicolumn{1}{l}{0.53 \tiny{$\pm$ 0.04}} & \multicolumn{1}{l}{0.69 \tiny{$\pm$ 0.05}} \\
\emph{L$^3$ (Ours)}                     & \textbf{0.92 \tiny{$\pm$ 0.02}}                     & \textbf{0.95 \tiny{$\pm$ 0.03}}                     & \textbf{0.97 \tiny{$\pm$ 0.02}}                      & \textbf{0.93 \tiny{$\pm$ 0.02}}                     & \textbf{0.94 \tiny{$\pm$ 0.03}}                     & \textbf{0.98 \tiny{$\pm$ 0.02}}                     \\ \hlineB{3}
\end{tabular}
\end{adjustbox}
}
\end{minipage}
\begin{minipage}[c]{0.5\textwidth}
\centering
\label{table:domain_adapt}
\renewcommand{\arraystretch}{1.21}
{
\begin{adjustbox}{width=0.97\textwidth}
\setlength{\tabcolsep}{0.3em}
\begin{tabular}{cccc|ccc}
\hlineB{3}
Direction       & \multicolumn{3}{c|}{Real $\rightarrow$ Toy}           & \multicolumn{3}{c}{Real $\rightarrow$ Realistic}      \\ \hline
Models          & DCI $\uparrow$          & MIG $\uparrow$          & FVAE $\uparrow$         & DCI $\uparrow$          & MIG $\uparrow$          & FVAE $\uparrow$         \\ \hline
Baseline        & 0.19 \tiny{$\pm$ 0.02} & 0.18 \tiny{$\pm$ 0.03} & 0.35 \tiny{$\pm$ 0.03} & 0.24 \tiny{$\pm$ 0.03} & 0.22 \tiny{$\pm$ 0.07} & 0.35 \tiny{$\pm$ 0.03} \\
DANN            & 0.75 \tiny{$\pm$ 0.04} & 0.73 \tiny{$\pm$ 0.02} & 0.76 \tiny{$\pm$ 0.02} & 0.73 \tiny{$\pm$ 0.04} & 0.74 \tiny{$\pm$ 0.03} & 0.72 \tiny{$\pm$ 0.02} \\
ADDA            & 0.64 \tiny{$\pm$ 0.05} & 0.61 \tiny{$\pm$ 0.03} & 0.69 \tiny{$\pm$ 0.04} & 0.65 \tiny{$\pm$ 0.06} & 0.69 \tiny{$\pm$ 0.04} & 0.65 \tiny{$\pm$ 0.03} \\
\emph{L$^3$} (Ours) & \textbf{0.92 \tiny{$\pm$ 0.02}} & \textbf{0.95 \tiny{$\pm$ 0.03}} & \textbf{0.97 \tiny{$\pm$ 0.02}} & \textbf{0.93 \tiny{$\pm$ 0.02}} & \textbf{0.94 \tiny{$\pm$ 0.03}} & \textbf{0.98 \tiny{$\pm$ 0.02}} \\ \hlineB{3}
\end{tabular}
\end{adjustbox}
}
\end{minipage}
\end{table*}

%% file: sections/05_conclusion.tex
\section{Conclusion}
We proposed a novel framework, \emph{Look}, \emph{Learn} and \emph{Leverage} (\emph{L$^3$}), which focuses on improving the \emph{intrinsic relations} discovery model generalization. Each of the three phases has a distinct focus, where, by utilizing the class-agnostic segmentation masks as the common symbolic space, the pretrained relations discovery model can be directly employed on a new visual domain without requiring \emph{intrinsic relations} data. Outstanding performance on three tasks, DRL, CRL and VQA, along with comprehensive discussions demonstrate the advantage and potential of \emph{L$^3$}. 

%% file: sections/xx_appendix.tex
\newpage
\appendix
\onecolumn
\section{Appendix}

\subsection{Datasets Details}
\label{sec:dataset-details}

\subsubsection{CRL datasets generation}
As mentioned in \cref{sec:experiments}, in order to generate the CRL dataset in different domains, we utilize the Shadow-Datasets Blender file supplied by \cite{shadows} and adjust the materials of both object and the floor. By only changing the materials of object and floor, the underlying causal mechanisms of Shadow-Datasets stay unchanged. The comparison between the original Shadow-Dataset and the newly generated Shadow-Datasets: Metal can be illustrated in \cref{fig:sunlight-dataset-comparision,fig:pointlight-dataset-comparision}. 

To properly evaluate CRL performance, we utilize metrics proposed in \cite{do-VAE}. Even though Maximum information coefficients (MIC) and total information coefficients (TIC) \cite{mic-tic} are first proposed to be metrics for CRL by \cite{causalvae}, as analyzed by \cite{do-VAE}, MIC and TIC only measure the mutual information between a latent factor and the respected generative factor, where no causal relation is evaluated. Thus, new metrics PosMIC/TIC and NegMIC/TIC \cite{do-VAE} are proposed to evaluate the correctness and falseness of learned causal relationships, respectively. Specifically, to calculate PosMIC/TIC, the latent effect factors before the causal discovery layer (DECI \cite{deci} in our implementation), are first set to zero. Then, the new latent factors are propagated through the learned causal discovery layer, and the latent effect factors after the causal discovery layer are used to calculate MIC and TIC values with the ground-truth generative effect factors. If the correct relationships are learned by the causal discovery layer, the corrected value of effect factors can be successfully inferred from their corresponding cause factors. Therefore, higher values of PosMIC/TIC show better performance. On the contrary, to calculate NegMIC/TIC, the latent cause factors before the causal discovery layer are first set to zero. Then, after propagating latent factors, the MIC/TIC value between the ground-truth cause factors and latent cause factors after the causal discovery layer is assessed to be the final score. Since causal relationships should only propagate from causes to effects \cite{book-causality}, smaller values of NegMIC/TIC indicate better performance. In addition, to consider both positive and negative metrics altogether, $F_1^{MIC}$ and $F_1^{TIC}$ are obtained by calculating the harmonic mean between the positive metrics and one minus negative metrics. All metrics range from $0$ to $1$. Since $F_1^{MIC}$ and $F_1^{TIC}$ can reveal the overall performance, we include the performance of those two metrics in the main pages.

\subsubsection{VQA}
As mentioned in \Cref{sec:experiment_details}, VQA is conducted by using VQAv2~\cite{vqa_v2} dataset as source domain and TDW-VQA~\cite{simvqa} dataset as target domain due to TDW-VQA does not have sufficient data for solely training. We focused on the counting data for better matching the question type between two datasets. We utilize the counting question split file provide by TDW-VQA and filter the VQAv2 dataset by using the \emph{numeric} questions that include the keyword ``how'' in the question type. Further, following ~\cite{mcan,simvqa}, we use Faster RCNN~\cite{faster-rcnn} based bottom-up-attention~\cite{botton-up-att}, pretrained on Visual Genome~\cite{vgenome}, to extract visual features, which serves as the visual input $X$. For each image, a 100 $\times$ 2048 feature is extracted where each of the 2048 features represent the feature of a region proposal and the maximum number of features is 100. In the case that the number of feature is less than 100, zero features are added to unified the feature size to 100 $\times$ 2048. Since the visual feature, other than RGB image, is served as visual input, the feature extraction network can act as a distribution aligner, where, despite the quality of the output feature in different visual domains, they are likely belonging to similar distribution. Thus, to introduce noise to the input feature which mimic the scenario of visual domain shift, we take inspiration from DDIM~\cite{ddim}, and introduce noise to the visual feature by various diffusion timesteps. We studied the effect of adding [0, 50, 100, 200, 300] timesteps where the distribution will be degenerated to $\mathcal{N}(0,\,1)$ at 1000-th timestep. We split the TDW-VQA dataset in the ratio of 0.2:0.8 for \emph{leverage} alignment and test.

\subsection{Experiments Details}
\label{sec:experiments-details}

\subsubsection{Implementation Details of DRL}
In this section, we give a more detailed introduction to the implementation of AdaVAE \cite{weakly-disentanglement}, which we used as a downstream module in our work for the DRL task. To implement such a method, a pair of inputs needs to be generated. To create the pair of inputs, one sample is randomly chosen from the dataset first and another sample is strategically to make sure there is $k$ number of different generative factors between two samples in a pair. In our experiments, following the best empirical result setting in the original paper \cite{weakly-disentanglement}, we choose $k$ to be two. During the source-train stage, since a pair of inputs $x_s$ and $x'_s$ is taken as input, two corresponding latent factors $z_s$ and $z'_s$ are generated through the feature extractor. For those two latent factors, AdaVAE first calcualtes the KL divergence between each factor of such latent factor pair on the same location. Then, two locations that have the two biggest KL divergence values are chosen as the location of latent factors corresponding to the different generative factors. Further, a new pair of latent factors $\hat z_s$ and $\hat z'_s$ are generated by calculating the average value between the original latent factor pair on all locations except the two locations which have the biggest KL divergence value. The newly obtained latent factors $\hat z_s$ and $\hat z'_s$ are then fed into the decoder to generate two new reconstructions $\hat x_s$ and $\hat x'_s$. By comparing the new reconstructions with original inputs $x$ and $x'$, respectively, the  \emph{intrinsic relation} information existing in a strategically chosen pair is utilized. For more detail, we still refer readers to the original paper. In addition, during the target-adaptation stage, besides utilizing $F_{MP}$ and common symbolic space to align the visual feature $v_t$ to the original visual feature, we add another KL-divergence alignment loss between the feature $z_t$ and the feature transformed from $v_t$ by a linear layer. Since AdaVAE needs to use a reparameterization trick on the latent feature to make $z$ to be a normal distribution, the pre-trained downstream task module can serve as an additional anchor for adaptation by using such KL-divergence alignment loss.
We used Adam optimizer with $\beta=(0.9, 0.999)$. The batch size is set to 64 and the model is trained for 100 epochs for both source-train and target-adapt stages. For \emph{toy} source domain, the learning rates for the source-train and target-adapt stage are both 3e-4. For \emph{realistic} source domain, the learning rates for the source-train and target-adapt stage are both 7e-4. For \emph{real} source domain, the learning rates for the source-train and target-adapt stage are both 5e-4.

\subsubsection{Implementation Details of CRL}
As mentioned in \cref{sec:task-application}, we utilize CausalVAE as the \emph{intrinsic relation} discovery module for the CRL task. To train CausalVAE, the ground-truth generative factors are required and utilized in order to learn semantic factors and causal structures among those factors. As discussed in \cite{do-VAE}, CausalVAE is limited by the linear causal discovery layer utilized in its original implementation \cite{causalvae}. Therefore, to fully exceed the CausalVAE performance under a fully supervised setting, we utilize DECI \cite{deci} to replace the original linear causal discovery layer. We used Adam as the optimizer with $\beta=(0.9,0.999)$. The batch size is set to 64 and the model is trained for 100 epochs for both source-train and target-adapt stages. Learning rates for training on both source domains (\emph{original} and \emph{metal}) and adapting on another domain are 1e-4.

\subsubsection{Implementation Details of VQA}
As mentioned in \cref{sec:task-application}, we utilize MCAN as the downstream module for VQA task. Different with DRL and CRL where the final feature is fed to a VAE encoder for $z$, we simply use a feed-forward network to directly map $a^i$ to $\hat{a}^i$ to avoid information bottleneck effect. Then, the $\hat{a}^i$ will be used as both the visual input feature to MCAN and the input to reconstruct $\tilde{x}^i$ and $\tilde{m}^i$. The question as answer will serve as $Y_s$ and is fed to MCAN for VQA training \cite{mcan,simvqa}. We used Adam optimizer with $\beta=(0.9, 0.98)$. For the source training, the batch size is 64, the learning rate is 1e-4, and the model is trained for 13 epochs. For the target adapt, the batch size is 32, the learning rate is 1e-3, and the model is trained for 13 epochs.

\subsubsection{Computation Resources}
Experiments on DRL and CRL are conducted on one NVIDIA 1080Ti GPU, and experiments on VQA are conducted on one or two NVIDIA RTX8000 GPUs due to the complexity of the VQA task. Compared with recent ML state-of-the-art models that may even trained on several NVIDIA A100 GPUs, we believe \emph{$L^3$} is arguably not computationally demanding.

\subsection{Comparing with Conventional Domain Adaptation}
Conventional domain adaptation (DA) research has been conducted in the literature on \emph{visual appearance} recognition tasks (e.g., image classification, object detection, and semantic segmentation) with many data availability assumptions, such as unsupervised DA, source-free DA, and unsupervised source-free DA. Despite the effort to increase the adaptation difficulty, due to the nature of the appearance recognition tasks, the target raw visual input itself contains sufficient information for training. For example, in the image classification task, on the target domain, the image itself contains the object of interest (e.g., an image of a dog containing a dog) regardless of the availability of the image label. So that many unsupervised or self-supervised methods can still be applied to extract features regarding the object of interest. However, for discovering the \emph{intrinsic relations}, the training data needs to contain both objects \emph{visual appearance} data and also \emph{intrinsic relations} data. Without either data, end-to-end \emph{intrinsic relation} discovery models can not be trained or fine-tuned, which challenges the conventional domain adaptation approaches. Furthermore, in our problem formulation, source domain data is also inaccessible on the target domain, which further limits the applicability of conventional DA models that focus on visual appearance tasks. Although there are several works that also study the DA problem on \emph{intrinsic relations } discovery tasks, they also usually assume the availability of either target label data and/or the entire source domain data. We provide a table comparison in \cref{tab:setup_compare_da} to help readers better understand the challenge of our problem formulation and distinguish the difference compared to other DA problems. For better demonstration, we assume that, for the visual appearance recognition tasks, the visual appearance data itself contains intrinsic relation data due to the nature of the task. 

Furthermore, to the best of our knowledge, there is no suitable SOTA that focuses on the three discussed downstream tasks with the same data assumption, which can be directly used as the baseline. Also, many recent SOTAs are specifically crafted for a certain visual appearance recognition task, such as image classification, which is hard to be adapted for our usage. Thus, to complete the discussion and provide quantitative studies, we increase the data assumption strength by including source data for adaptation and adapt two more generalized approaches, DANN and ADDA, as the baseline. The results and relevant discussions are included in \cref{tab:ablation_fusion_adapt} and \cref{sec:discussion}.

\input{tables-latex/setup_compare_with_da}

\subsection{Broader Impacts, Limitations, and Future Works}
In this paper, we proposed a novel framework \emph{L$^3$} that aims to resolve the visual domain shift of \emph{intrinsic relations} discovery via symbolic alignment. Further, \emph{L$^3$} is also capable of utilizing extra auxiliary data to enhance or expand the \emph{intrinsic relations} discovery spectrum. Even in the challenging scenario where such auxiliary data is visual-domain shift vulnerable (e.g., RawVis) and \emph{intrinsic relations} data on the target domain is absent, our proposed alignment strategy also empowers the \emph{Look} module of \emph{L$^3$} to be adjusted on the target domain with only \emph{visual appearance} data where the pre-trained \emph{intrinsic relations} discovery module can be directly reused. Thus, one of the most promising positive broader impacts is that for certain \emph{intrinsic relations} discovery tasks where the data collection is challenging in the real world, comprehensive (appearance and relations) training data can be generated in the synthetic domain to train the entire \emph{L$^3$}, and only \emph{visual appearance} data is needed on the target domain for the adaptation (if needed as discussed in \cref{sec:discussion}). For example, \emph{intrinsic relations} of the accident and natural disaster-related tasks are hard to collect, and \emph{intrinsic relations} of diseases are not only hard to collect but also may not be ethical to collect. Thus, \emph{L$^3$} has huge potential to address the challenges in those scenarios. Furthermore, we also see other positive broader impacts of \emph{L$^3$}, for example, ease the data collection process and free from any duplicated training for the cases that, despite the visual domain being different, the underlying \emph{intrinsic relations} and mechanisms are the same.

We also acknowledge that our work has limitations and opens several possible future works. In the paper, to realize \emph{L${^3}$} while intentionally maintaining a limited data requirement strength to demonstrate the generality, effectiveness, and robustness of the framework, we select class-agnostic SegMasks as the primary symbolic space and RawVis as the auxiliary data. It may be arguable that class-agnostic SegMasks are not comfortable fitting in a narrow definition of symbolic space because the masks lack semantic meaning. However, we want to highlight that the class-agnostic SegMasks are still human-interpretable compared to high-dimensional features (as shown in \cref{fig:mpi3d-visualization}). Thus, we see the class-agnostic SegMasks still fall into a broader definition of symbolic space. Of course, we see that there are many other possible selections, such as semantic SegMasks, instance SegMasks, object bounding boxes, human skeletons, etc. Although we are not able to exclusively list and evaluate each of the possible selections, we hope our work can serve as a basic and generalized starting point that can inspire future works to adopt, customize, and even simplify \emph{L${^3}$} according to various \emph{intrinsic relations} discovery tasks and data assumption strength. Additionally, we also find that, compared to \emph{visual appearance} recognition tasks, collecting and pairing datasets for \emph{intrinsic relations} discovery tasks, where the visual domain is different while the underlying \emph{intrinsic relations} are the same, is considerably challenging. This challenge may largely limit the scope of studies (e.g., small dataset scales, limited dataset tasks, and missing real-world data). Thus, we also hope our work can motivate future works and draw more attention to this challenging problem so that the community can collaboratively create large-scope and comprehensive datasets for various \emph{intrinsic relations} discovery tasks.

\subsection{Unnormalized Experiment Results}
As mentioned in \cref{sec:experiments}, we report normalized results for DRL and CRL. In this section, we report the original results value of DRL and CRL in \cref{fig:mpi3d-total-results-unnormalized} and \cref{table:causal_results_unnormalized}, respectively.
\input{figures-latex/sunlight_dataset_comparision}
\input{figures-latex/pointlight_dataset_comparision}
\input{figures-latex/mpi3d_results_unnormalized}
\input{figures-latex/mpi3d_source_results}
\input{tables-latex/causal_results_unnormalized}
\input{tables-latex/causal_results_source}

%% file: tables-latex/setup_compare_with_da.tex
\begin{table}[ht]
\centering
\label{tab:setup_compare_da}
\caption{Target domain data availability comparison between our problem formulation and several domain adaptation problems that primarily focus on recognizing object visual appearance.}
\renewcommand{\arraystretch}{1.21}
{
\begin{adjustbox}{width=0.95\linewidth}
\begin{tabular}{ccccc}
\hlineB{3}
Target Data Availibility    & Source Domain Data    & Target Task Ground-truth     & Visual Appearance Data & Intrinsic Relations Data \\ \hline
Supervised DA               & \cmark & \cmark & \cmark  & \cmark     \\
Unsupervised DA             & \cmark & \xmark & \cmark  & \cmark     \\
Source-Free DA              & \xmark & \cmark & \cmark  & \cmark     \\
Unsupervised Source-Free DA & \xmark & \xmark & \cmark  & \cmark     \\
Our Problem Formulation     & \xmark & \xmark & \cmark  & \xmark     \\ \hlineB{3}
\end{tabular}
\end{adjustbox}
}
\end{table}

%% file: figures-latex/sunlight_dataset_comparision.tex
\begin{figure*}
    \centering
     \includegraphics[width=0.95\textwidth]{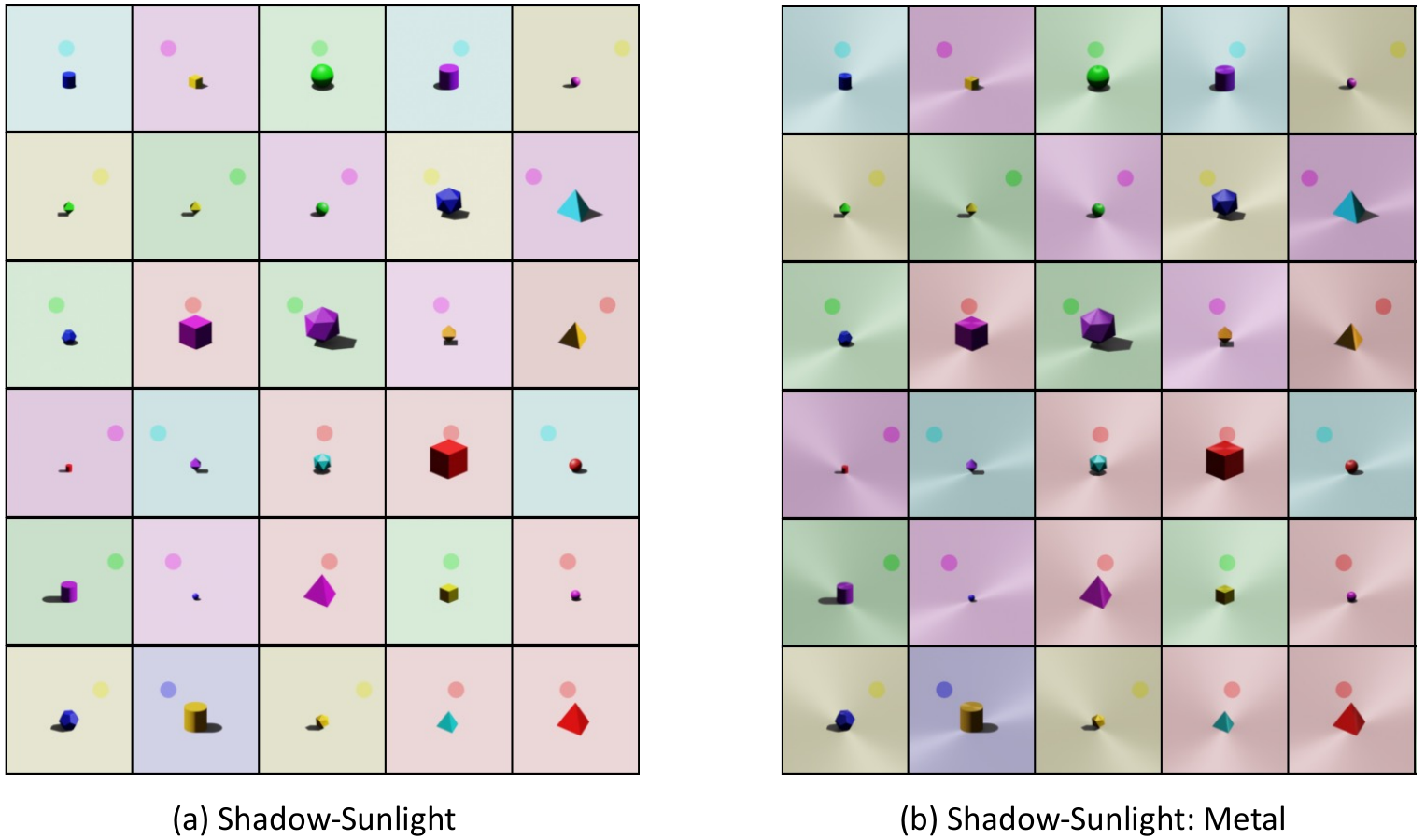}
    \caption{Comparison between original Shadow-Sunlight and Shadow-Sunlight: Metal, which is generated by us.} %
    \label{fig:sunlight-dataset-comparision}
\end{figure*}

%% file: figures-latex/pointlight_dataset_comparision.tex
\begin{figure*}
    \centering
     \includegraphics[width=0.92\textwidth]{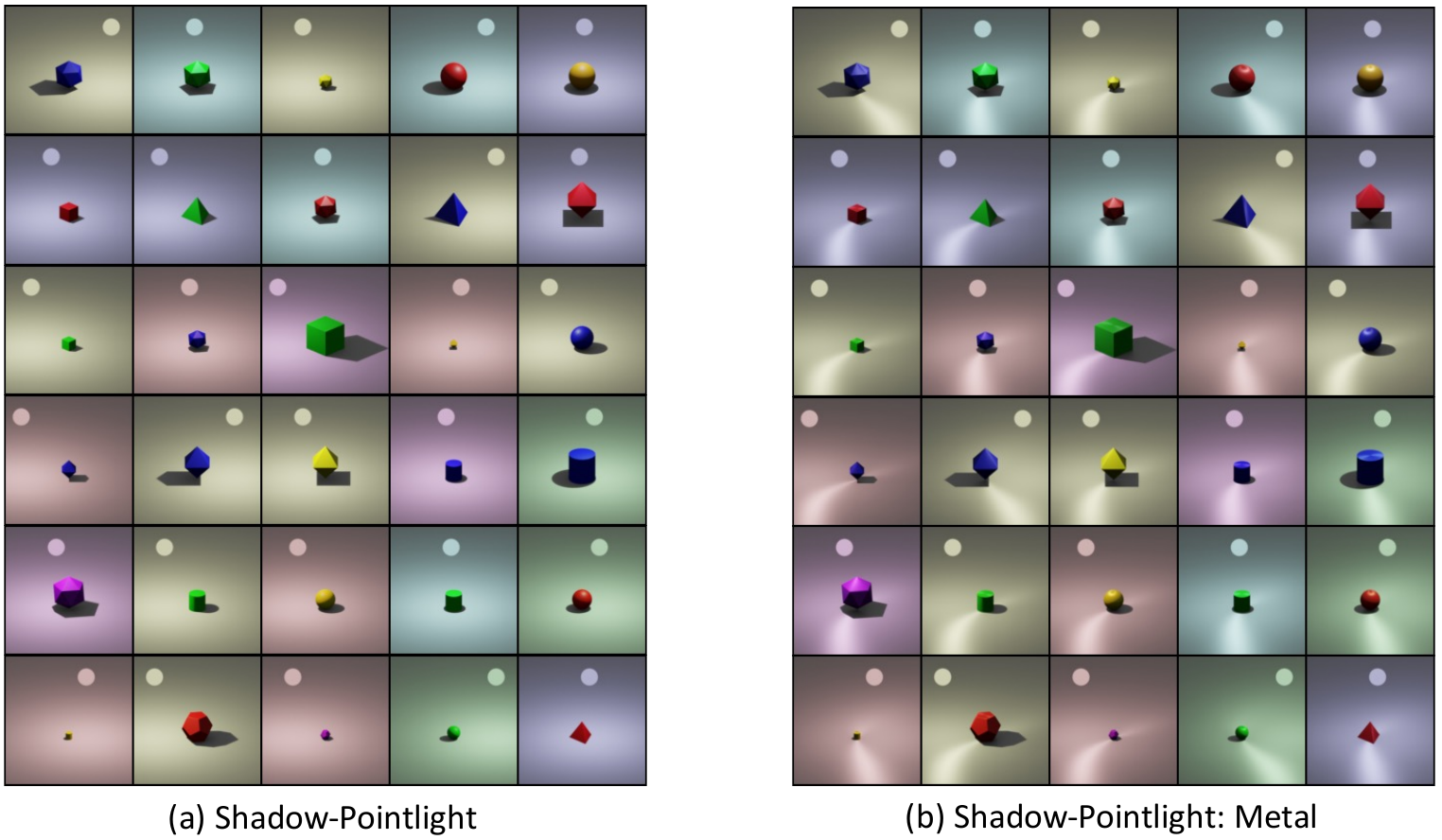}
    \caption{Comparison between original Shadow-pointlight and Shadow-pointlight: Metal, which is generated by us.} %
    \label{fig:pointlight-dataset-comparision}
\end{figure*}

%% file: figures-latex/mpi3d_results_unnormalized.tex
\begin{figure*}
    \centering
     \includegraphics[width=0.85\textwidth]{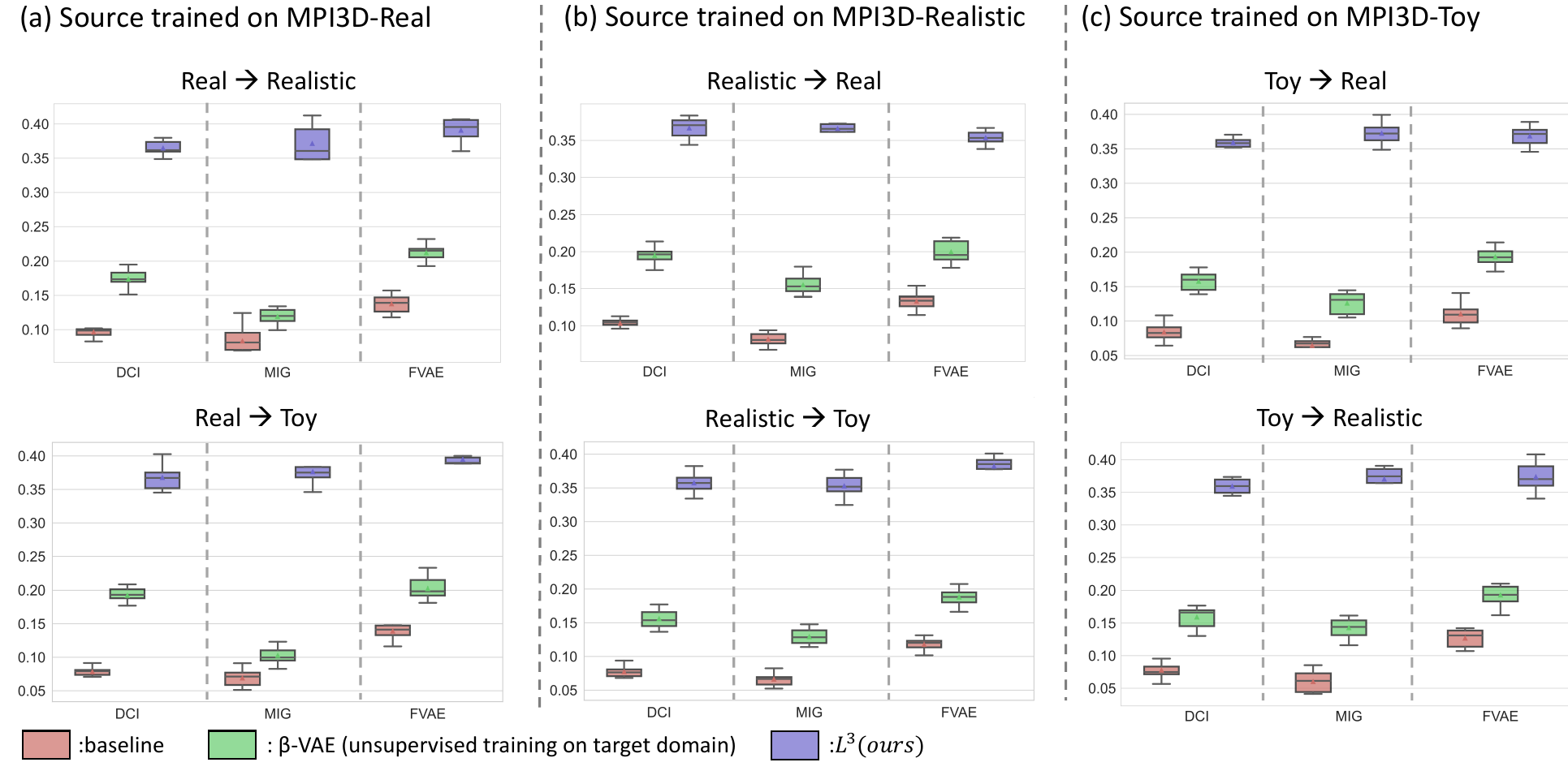}
    \caption{Performance of DRL task on MPI3D dataset. Unnormalized value results of each metric are reported.} %
    \label{fig:mpi3d-total-results-unnormalized}
\end{figure*}

%% file: figures-latex/mpi3d_source_results.tex
\begin{figure*}
    \centering
     \includegraphics[width=0.85\textwidth]{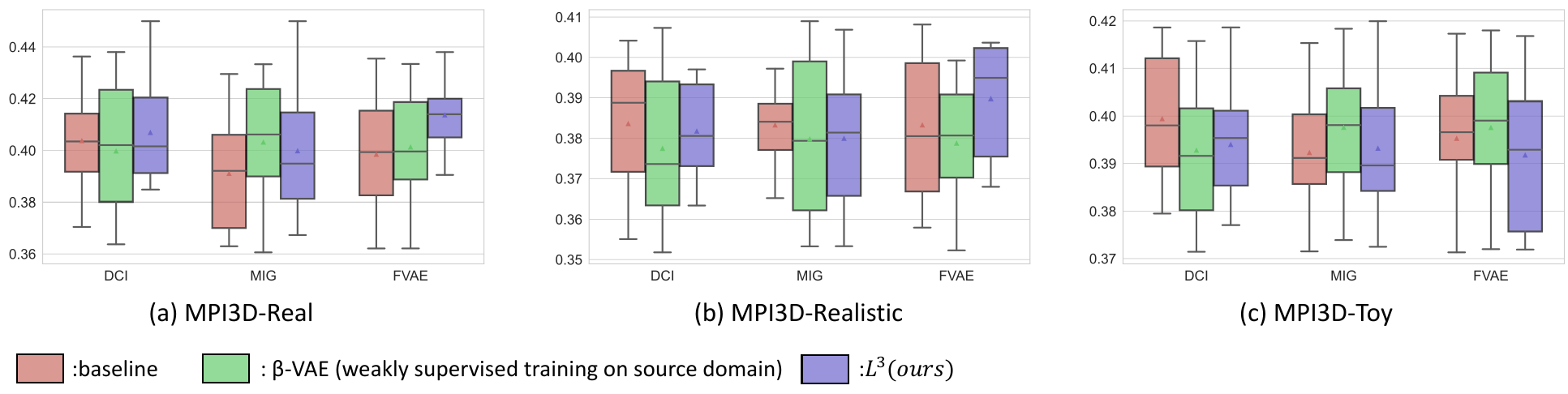}
    \caption{Performance of DRL on source domain of each MPI3D subsets. UT is trained in the weakly supervised setting in the source domain and will be trained in an unsupervised learning setting in the target domain.} %
    \label{fig:mpi3d-total-results-source}
\end{figure*}

%% file: tables-latex/causal_results_unnormalized.tex
\begin{table*}[ht]
\centering
\caption{Causal Representation Learning result on the Shadow Dataset. Original result is reported.}
\label{table:causal_results_unnormalized}
\renewcommand{\arraystretch}{1.2}
{
\begin{adjustbox}{width=0.9\textwidth}
\setlength{\tabcolsep}{0.3em}
\begin{tabular}{ccccc|cccc}
\hlineB{3}
Dataset   & \multicolumn{4}{c|}{Shadow-Sunlight}                                      & \multicolumn{4}{c}{Shadow-Pointlight}                                    \\ \hline
Direction & \multicolumn{2}{c|}{Original $\rightarrow$ Metal}   & \multicolumn{2}{c|}{Metal $\rightarrow$ Original} & \multicolumn{2}{c|}{Original $\rightarrow$ Metal}   & \multicolumn{2}{c}{Metal $\rightarrow$ Original} \\ \hline
Models    & $F_1^{MIC}\uparrow$ & \multicolumn{1}{c|}{$F_1^{TIC}\uparrow$ } & $F_1^{MIC}\uparrow$           & $F_1^{TIC}\uparrow$          & $F_1^{MIC}\uparrow$ & \multicolumn{1}{c|}{ $F_1^{TIC}\uparrow$ } & $F_1^{MIC}\uparrow$          & $F_1^{TIC}\uparrow$          \\ \hline
Baseline  & 0.59 $\pm$ 0.04   & \multicolumn{1}{c|}{0.60 $\pm$ 0.03}   & 0.64 $\pm$ 0.03             & 0.65 $\pm$ 0.03           & 0.63 $\pm$ 0.04   & \multicolumn{1}{c|}{0.61 $\pm$ 0.05}   & 0.65 $\pm$ 0.02            & 0.69 $\pm$ 0.03           \\
UT        & 0.36 $\pm$ 0.03   & \multicolumn{1}{c|}{0.32 $\pm$ 0.04}   & 0.33 $\pm$ 0.02             & 0.35 $\pm$ 0.03            & 0.34 $\pm$ 0.04   & \multicolumn{1}{c|}{0.33 $\pm$ 0.04}   & 0.36 $\pm$ 0.03            & 0.38 $\pm$ 0.03            \\
\emph{L$^3$} (Ours)  & \textbf{0.85 $\pm$ 0.03}   & \multicolumn{1}{c|}{\textbf{0.86$\pm$ 0.03}}   & \textbf{0.85 $\pm$ 0.02}             & \textbf{0.86 $\pm$ 0.03}            & \textbf{0.85  $\pm$ 0.02}   & \multicolumn{1}{c|}{\textbf{0.84 $\pm$ 0.02}}   & \textbf{0.85 $\pm$ 0.02}            & \textbf{0.84 $\pm$ 0.02}            \\ \hlineB{3}
\end{tabular}
\end{adjustbox}
}
\end{table*}

%% file: tables-latex/causal_results_source.tex
\begin{table*}[ht]
\centering
\caption{Causal Representation Learning result on the source of Shadow Dataset.  UT is trained in the fully supervised setting in the source domain and will be trained in an unsupervised learning setting in the target domain. }
\label{table:causal_results_source}
\renewcommand{\arraystretch}{1.2}
{
\begin{adjustbox}{width=0.9\textwidth}
\setlength{\tabcolsep}{0.3em}
\begin{tabular}{ccccc|cccc}
\hlineB{3}
Dataset   & \multicolumn{4}{c|}{Shadow-Sunlight}                                      & \multicolumn{4}{c}{Shadow-Pointlight}                                    \\ \hline
Direction & \multicolumn{2}{c|}{Original}   & \multicolumn{2}{c|}{Metal } & \multicolumn{2}{c|}{Original }   & \multicolumn{2}{c}{Metal} \\ \hline
Models    & $F_1^{MIC}\uparrow$ & \multicolumn{1}{c|}{$F_1^{TIC}\uparrow$ } & $F_1^{MIC}\uparrow$           & $F_1^{TIC}\uparrow$          & $F_1^{MIC}\uparrow$ & \multicolumn{1}{c|}{ $F_1^{TIC}\uparrow$ } & $F_1^{MIC}\uparrow$          & $F_1^{TIC}\uparrow$          \\ \hline
Baseline  & 0.89 $\pm$ 0.04   & \multicolumn{1}{c|}{0.87 $\pm$ 0.03}   & 0.87 $\pm$ 0.03             & 0.88 $\pm$ 0.03           & 0.87 $\pm$ 0.04   & \multicolumn{1}{c|}{0.88 $\pm$ 0.04}   & 0.89 $\pm$ 0.02            & 0.89 $\pm$ 0.03           \\
UT        & 0.87 $\pm$ 0.03   & \multicolumn{1}{c|}{0.86 $\pm$ 0.04}   & 0.86 $\pm$ 0.02             & 0.87 $\pm$ 0.03            & 0.86 $\pm$ 0.04   & \multicolumn{1}{c|}{0.87 $\pm$ 0.04}   & 0.88 $\pm$ 0.03            & 0.87 $\pm$ 0.03            \\
\emph{L$^3$} (Ours)  & \textbf{0.89 $\pm$ 0.03}   & \multicolumn{1}{c|}{\textbf{0.89$\pm$ 0.03}}   & \textbf{0.87 $\pm$ 0.02}             & \textbf{0.87 $\pm$ 0.03}            & \textbf{0.88  $\pm$ 0.02}   & \multicolumn{1}{c|}{\textbf{0.84 $\pm$ 0.02}}   & \textbf{0.87 $\pm$ 0.02}            & \textbf{0.84 $\pm$ 0.02}            \\ \hlineB{3}
\end{tabular}
\end{adjustbox}
}
\end{table*}